\documentclass[cameraready]{Interspeech}

\usepackage{cite}
\usepackage{multirow}     
\usepackage{tabularx}     
\usepackage{colortbl}     

\ifdefined\pdfobjcompresslevel \pdfobjcompresslevel=0 \fi

\newcommand{\bengsnip}{\raisebox{-0.15ex}{\includegraphics[height=1.5ex]{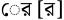}}}

\title{Dynamic Block-Online Streaming ASR for Low-Resource Agglutinative Code-Switching Speech with Morphology-Aware Evaluation}

\author[affiliation={1},correspondingauthor]{Kazi}{Rafat}
\author[affiliation={1}]{Afifa}{Imran}
\author[affiliation={1}, orcid=0000-0002-4665-0530,]{Md. Ismail}{Hossain}
\author[affiliation={1}]{Md. Romzan}{Ali}
\author[affiliation={2}, orcid=0000-0002-8670-7124,]{Fuad}{Rahman}
\author[affiliation={1}, orcid=0000-0001-8683-7247,]{Sifat}{Momen}
\author[affiliation={1}, orcid=0000-0001-7169-0318,]{Shafin}{Rahman}
\author[affiliation={1}, orcid=0000-0002-7661-3570,]{Nabeel}{Mohammed}

\address{
    $^1$ Address Affiliation 1, Country Affiliation 1 \\
    $^2$ Address Affiliation 2, Country Affiliation 2 \\
    $^3$ Address Affiliation 3, Country Affiliation 3
}
\address{
    $^1$ Apurba NSU R\&D Lab, North South University, Dhaka, Bangladesh \\
    $^2$ Apurba Technologies, Dhaka, Bangladesh }

\email{kazi.meem@northsouth.edu, afifa.imran@northsouth.edu, ismail.hossain2018@northsouth.edu, romzanalinsu@gmail.com, fuad@apurbatech.com, sifat.momen@northsouth.edu, shafin109@gmail.com, nabeel.mohammed@northsouth.edu}


\usepackage{comment}

\begin{document}

\maketitle

\begin{abstract}
Low-resource, agglutinative languages challenge low-latency ASR, as fixed lookahead windows fail when grapheme order diverges, suffixes cross code-switch boundaries, or code-switching data is imbalanced. We present the first comprehensive study of streaming ASR for intra-sentential Bangla–English code-switching, identifying a mismatch between strict causal streaming and long-range morphological dependencies. We introduce a Dynamic Block-Online framework\footnote{Code, metric and lexicon are available online: \href{https://github.com/King-Rafat/Dynamic_Streaming_ASR}{Dynamic ASR}} with VAD-aligned inference that balances semantics, latency, and context, enabling global bidirectional attention to preserve morphological integrity. To mitigate data scarcity, we propose Script-Anchored Loanword Injection, augmenting training text with English loanwords to model phonotactic transitions. We introduce CS-WER to diagnose switch, root, and suffix errors, and show that flexible latency budgets enhance recognition of agglutinative and mixed-language speech, with successful transfer to menstrual-medical domains.
\end{abstract}

\section{Introduction}
\label{sec:introduction}
Recent ASR systems are multilingual~\cite{radford2023robust}, realtime~\cite{graves2012sequence}, and domain-adaptable. This progress is largely driven by Large Language Models (LLMs), which necessitate low-latency ASR front-ends for real-time interaction. Recent advancements in ASR have been driven by scale, utilizing thousands of hours of training data and costly compute~\cite{narayanan2018toward, sainath2020streaming}. Streaming reaches low-resource Indic languages mainly via monolingual transducer/CTC models; intra-sentential code-switching in agglutinative scripts remains largely unaddressed, particularly within the constraints of low-latency architectures~\cite{gao2022paraformer, li2021better}.

Online systems enable real-time correction and reduce SLM latency through incremental token output. However, low-resource ASR systems are rarely public~\cite{kannan2019large} and hard to reproduce, due to scarce data and recipes. Although Bengali is encoded in phonetic order, dependent vowel signs render left of their consonant. Causal streaming must emit the consonant first, giving unstable output, worst at switch boundaries. Colloquial Bengali incorporates English loanwords, requiring robust handling of \textbf{intra-sentential code-switching (CS)}. Accurate CS recognition is crucial for modern AI pipelines~\cite{sitaram2019survey} because preserving high-information loanwords minimizes semantic drift in downstream tasks.

\begin{figure}[!t]
    \centering
    \includegraphics[scale=0.9]{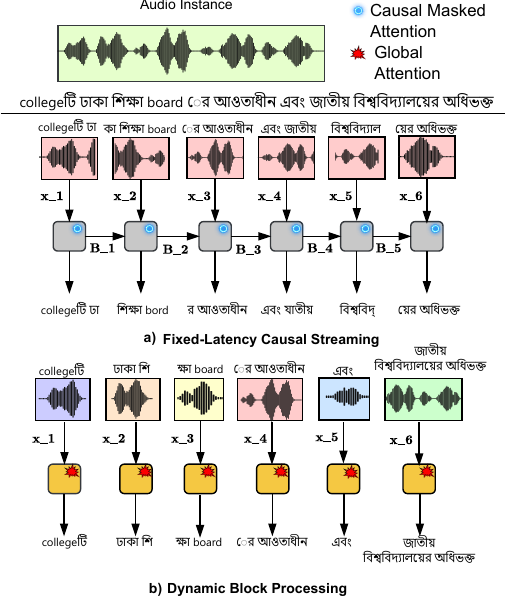}
    \caption{Comparison of (a) Fixed-Latency Causal Streaming, where memory-cached attention leads to CS morphological truncation (mistranscribing \bengsnip and board [bord]), and (b) our proposed Dynamic Block-Processing, where VAD-aligned global attention resolves long-form agglutinative dependencies.}
    \label{fig:D1}
\end{figure}

This work presents a study of streaming ASR for agglutinative CS in a low-resource setting. We show that standard causal models face limitations because linguistic units such as pre-base vowels, Bangla suffixes attached to English roots, and code switches often extend beyond the model’s lookahead window. In such cases, strictly causal attention fails to resolve these dependencies, as it lacks the hindsight context necessary to re-evaluate roots in light of future suffixes. Parallel to the architectural challenge, we address the scarcity of training data for CS. Colloquial Bangla is heavily interspersed with English loanwords. We introduce Script-Anchored Loanword Injection, embedding 750 common English loanwords into training text to teach phonotactic transitions and establish a strong “Banglish” baseline absent from open datasets.

We address the architectural mismatch with a Dynamic Block-Online architecture. Unlike fixed-window streaming that arbitrarily truncates morphemic suffixes, it uses Voice Activity Detection to align inference with natural pauses, producing dynamic latency—often faster than fixed buffers during rapid turn-taking, but extending to a 3-second macro-block. Within each block, an offline model trained with global bidirectional attention resolves morphological dependencies. This shows that for agglutinative languages, a flexible latency budget can compensate for scarce data and compute with an optimizable inference context, while preserving real-time capability. Figure~\ref{fig:D1} shows a model trained with global attention inferred on pauses and macro-blocks is able to better understand switches and suffixes as opposed to causal-masked streaming model. 
We further validate our approach via domain adaptation on a specialized menstrual and menopausal health dataset. Our architecture preserves morphological and CS-semantic integrity, so technical medical terms remain intact for downstream small language models.


\noindent Our key contributions are as follows:
\begin{itemize}
    \item \textbf{Dynamic Block-Online Processing:} A VAD-triggered streaming strategy for offline global attention models, optimized for agglutinative CS speech.
    \item \textbf{Script-Anchored Injection:} A novel synthetic augmentation technique using curated loanword datasets to induce intra-sentential CS in colloquial monolingual corpora.
    \item \textbf{CS-WER Metric:} A fine-grained evaluation framework to quantify error rates specifically across CS-roots and agglutinative suffixes.
    \item \textbf{Domain Generalization:} Validation of model robustness in the specialized health domain, specifically menstrual and menopausal health.
\end{itemize}

\section{Streaming Agglutinative Code-Switched (CS) ASR}
\subsection{Script-Anchored Synthetic CS Induction}
CS requires\textit{ vast good quality} data. Massively multilingual models~\cite{radford2023robust} perform well on inter-sentential multilingual speech but show limitations in intra-sentential CS scenarios~\cite{aditya2024attention}.

We define the monolingual Bengali corpus $D_{BN} = \{d_1,\dots,d_n\}$, where $d_i$ represents an utterance consisting of a token sequence, $\{\bm{w_1},\dots,\bm{w_k}\}$ from $V_{BN}$. We assume a morphological decomposition for each token $w = \text{root}(w) \oplus \text{suff}(w)$. The augmentation process is defined as a mapping lexicon $\mathcal{M} = \{(r_{bn}, r_{en}) \mid r_{bn} \equiv r_{en}\}$, consisting of 750 high-frequency semantic equivalents. The injection process identifies any Bengali token $w\in d_i$ if $\text{root}(w) \in \mathcal{M}$, which is transformed to its cs variant $w_{cs}$.
\begin{equation}
w_{cs} = r_{en} \oplus \text{suff}(w)
\end{equation}
Most synthetic methods mix semantically and acoustically mismatched audio, hindering convergence. In contrast, this approach preserves syntax while inserting English roots to create a code-switched corpus.











\begin{figure}[!t]
    \centering
    \includegraphics[scale = 1]{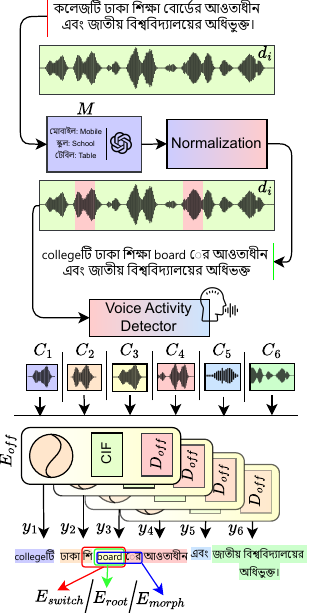}
    \caption{Proposed Methodology of: \textbf{Script-Anchored Loanword Injection} - converting Bengali to CS transcript through a loanword database $\mathcal{M}$. \textbf{A Dynamic Block Online model} that strips audio at communicative pauses and uses an offline encoder $E_{off}$, CIF predictor, and Decoder and how \textbf{CS-WER} is calculated showcased by what each of $E_{switch}/E_{root}/E_{morph}$ looks at.}
    \label{fig:D2} 
\end{figure}

\subsection{Dynamic Block and Streaming ASR}
RNN-T streaming models~\cite{graves2012sequence} are often sample-inefficient~\cite{karita2019comparative} and autoregressive approaches like Whisper \cite{radford2023robust} can produce hallucinations. To address these issues, we adopt a Non-Autoregressive (NAR) Paraformer \cite{gao2022paraformer} as our backbone that enables fast, low-compute NAR chunk decoding for stable reordering-heavy settings. The encoder $\mathcal{E}$ transforms an acoustic sequence $\mathbf{X} = \{\mathbf{x}_1, \dots, \mathbf{x}_T\}$, where $\mathbf{x}_t \in \mathbb{R}^f$ of utterance $d_i$, into a hidden representation $\mathbf{H} = \{\mathbf{h}_1, \dots, \mathbf{h}_T\}$.
$\mathcal{E}$ employs Self-Attention Network with Memory (SAN-M) blocks. During offline training, the architecture learns global bidirectional attention across an entire utterance.
In streaming, the architecture is trained by restricting the attention receptive fields. The representation for the frame $t$ is constrained to the current chunk $\mathbf{C}_k$ and a historical context memory bank $\mathbf{B}_{k-1}$: 
$\mathbf{h}_t = \text{SAN-M}(\mathbf{x}_t, \mathbf{B}_{k-1};\mathbf{C_k}) \quad \forall t \in \mathbf{C}_k$. 
This causality creates a future-blindness bottleneck, hindering the resolution of morphological boundaries in agglutinative scripts. To align $\mathbf{H}$ with target tokens $\mathbf{Y}$, a CIF mechanism emits weights $\alpha$, where $\alpha_t \in [0, 1]$. The acoustic embedding $\mathbf{c}_l$ for token $y_l$ is integrated as: $\mathbf{c}_l = \sum_{t=t_{l-1}}^{t_l} \alpha_t \mathbf{h}_t, \quad \text{s.t. } \sum_{i=1}^{t_l} \alpha_i \approx l$

 
We evaluate two streaming policies to investigate the impact of context restriction. \textit{\textbf{(1)} Causal Streaming ($\lambda \le 2.4s$):} Uses chunk-causal attention: full attention within a chunk, no access to future chunks. We posit that scaling $\lambda$ in this setup yields diminishing returns, as the model cannot look back from a suffix to correct an earlier root embedding, leading to irreversible alignment drift.
\textit{\textbf{(2)} Dynamic Block-Processing ($\lambda \le 3.0s$):} To mitigate this, we move to a VAD-triggered block-online paradigm. We substitute fixed causal updates with semantic macro-blocks aligned with natural communicative pauses. Crucially, this allows us to deploy the Unmasked Offline Encoder within the buffer. By treating the 3s buffer as a standalone unit, we restore Global Bidirectional Attention. This achieves hindsight resolution: the model re-evaluates early frames (roots) in light of later frames (suffixes) within the same block. This also helps to understand instances of code-switches. This eliminates the 'blind commitment' inherent in causal architectures, effectively utilizing the extra latency to enforce morphological integrity. Consequently, the effective receptive field expands to $\mathcal{R}(\mathbf{h}_t) = \{\mathbf{x}_i \mid \mathbf{x}_i \in \mathbf{C}_{block}\}$. covering the entire semantic unit regardless of where $t$ lies.

\subsection{Fine-Grained CS-WER}
Standard Word Error Rate (WER) treats all errors equally. A model might correctly recognize the English root but fail to segment the suffix, or miss the language switch entirely. To disentangle these errors, we introduce \textbf{Fine-Grained CS-WER}, a decomposed metric that evaluates three structural components of intra-sentential CS:
\begin{itemize}
\item \textbf{Switch-Point Error ($E_{switch}$):} Measures the stability of the language transition. We evaluate the bigram formed by the last word of the Matrix Language (Bangla) and the first word of the Embedded Language (English). 

\item \textbf{Loanword Root Error ($E_{root}$):} Isolates the lexical accuracy of the embedded English root, ignoring the suffix. This measures whether the acoustic model correctly identified the core English entity.

\item \textbf{Morphological Boundary Error ($E_{morph}$):} It evaluates the full agglutinative structure. A prediction is considered correct \textbf{only if} both the English root AND the Bangla suffix are correctly recognized and validly segmented. 
\end{itemize}
We show this visually at the end of Fig~\ref{fig:D2}. We report performance as a tuple $E_{switch}/E_{root}/E_{morph}$. This formulation allows us to specifically diagnose whether performance gains stem from sudden linguistic switches, better acoustic recognition, or superior morphological segmentation.
\section{Experiments}

\subsection{Experimental Setup}

\noindent\textbf{Datasets:}
We utilized a composite Bangla corpus of approximately 500 hours, aggregating the Common Voice Bengali split~\cite{alam2022bengali} ($\approx 75h$), OpenSLR 53~\cite{kjartansson-etal-sltu2018} ($\approx 215h$), IndicVoices~\cite{DBLP:conf/acl/JavedNGJBMSAFPR24} ($\approx 122h$), and KathBath~\cite{IndicSuperb} ($\approx 84h$). To ground the acoustic representation of English phonemes, we incorporated a 250h subset of English Gigaspeech~\cite{GigaSpeech2021} for approximately 750 hours in total. The Script-Anchored Loanword Injection resulted in $\approx 20\%$ CS sentences among all instances. 
\begin{table}[!t]
\centering
\small
\footnotesize 
\setlength{\tabcolsep}{2pt} 
\begin{tabularx}{\columnwidth}{|l|c|c|c|c|>{\centering\arraybackslash}X|}
\hline
 \multirow{2}{*}{Model}  & \multirow{2}{*}{Data} & \multirow{2}{*}{w} & \multicolumn{3}{c}{Common Voice}  \\
\cline{4-6}
& & & CER & WER & CS-WER  \\
\hline
\rowcolor{gray!20}
\hline
\multicolumn{6}{|l|}{Whisper Offline} \\
\hline
 Largev3~\cite{radford2023robust}   & -  & Full &  -  & 40.30 & .94/.96/.97 \\
  \hline
Largev2~\cite{radford2023robust}   & -  & Full & -  & 62.31 & .98/.94/.99 \\
 \hline
 \rowcolor{gray!20}
\multicolumn{6}{|l|}{Others Offline} \\
\hline
 MBNSpeech~\cite{nandi2023pseudo} & -  & Full & 11.44  & 42.3 & 1/1/1 \\
  \hline

 MMS~\cite{pratap2024scaling} & -  & Full & 14.72  & 48 & 1/1/1 \\
  \hline

 \hline
\rowcolor{gray!20}
\multicolumn{6}{|l|}{Paraformer Offline} \\
\hline
 + 750h \newline+ FC   & B+E  & Full &12.19  & 34.81 & .92/.87/.90 \\
 \hline
 + 750h \newline+ FC   & B+E+C& Full  & 12.69 & 33.32 & .53/.29/.36 \\
 \hline
 \hspace{0.5cm}+ aug & B+E+C & Full & \textbf{11.05} & \textbf{31.14} &  \textbf{.46/.25/.31} \\
\hline
\rowcolor{gray!20}
\multicolumn{6}{|l|}{Paraformer Streaming}  \\
\hline
+ 200h \newline   & B+E&600ms & 26.88 & 60.90 &  1/.99/1\\
\hline
+ 200h \newline  & B+E+C&600ms & 27.97& 61.44 & .72/.61/.67 \\
\hline
+ 200h \newline + LID & B+E+C&600ms & 33.45& 64.78  & .75/.67/.87 \\
\hline
+ 200h \newline + WLID   & B+E+C&600ms & 34.57& 65.18  & .75/.67/.87 \\
\hline
+ 750h \newline  & B+E+C&600ms & 22.96 & 51.72  & .69/.50/.53 \\
\hline
+ 750h \newline  & B+E+C&800ms & 21.41 & 49.43 & .68/.48/.55 \\
\hline
+ 750h \newline  & B+E+C& 2s & 20.19 & 48.88 & .66/.44/.51 \\
\hline
+ 750h \newline  & B+E+C& 2.4s & 18.17 & 47.45  &  .64/.42/.48\\
\hline

 \hspace{0.5cm}+ aug & B+E+C & 2s  &16.33  & 41.45 &  .62/.36/.44\\
 \hline
 \hspace{0.5cm}+ aug & B+E+C & 2.4s  &14.67  & 39.45 &  .60/.36/.43\\
 \hline
\hspace{0.5cm}+ aug & B+E+C & 3s  & \textbf{14.27}  & \textbf{37.20} &  .58/.35/.42\\
 \hline
\rowcolor{gray!20}
\multicolumn{6}{|l|}{Dynamic Block Paraformer}\\
\hline
+ 750h + aug & B+E+C & 3s& 15.62 & 38.73 & \textbf{.53/.29/.35}\\
\hline
\end{tabularx}
\caption{Comparison of CER, WER, and Fine-Grained CS-WER ($E_{switch}/E_{root}/E_{morph}$) across Offline, Causal Streaming, and Dynamic Block architectures. 'B+E', 'C', and 'aug' denote monolingual training, loanword injection, and noise augmentation, respectively.}
\label{tab:Results1}
\end{table}
\noindent\textbf{Test Sets:}
Evaluation is performed on two distinct domains: (1) a conversational test set (Common Voice, 9,392 utterances) with common English loanwords using Script-Anchored Injection, and (2) Medical Domain: 20 h high-perplexity dataset focusing on Menstrual Health for training and an additional 5 h for testing. 
\begin{figure*}[htbp]
    \centering
    \begin{minipage}{.9\textwidth}
        \centering
        \includegraphics[width=\linewidth]{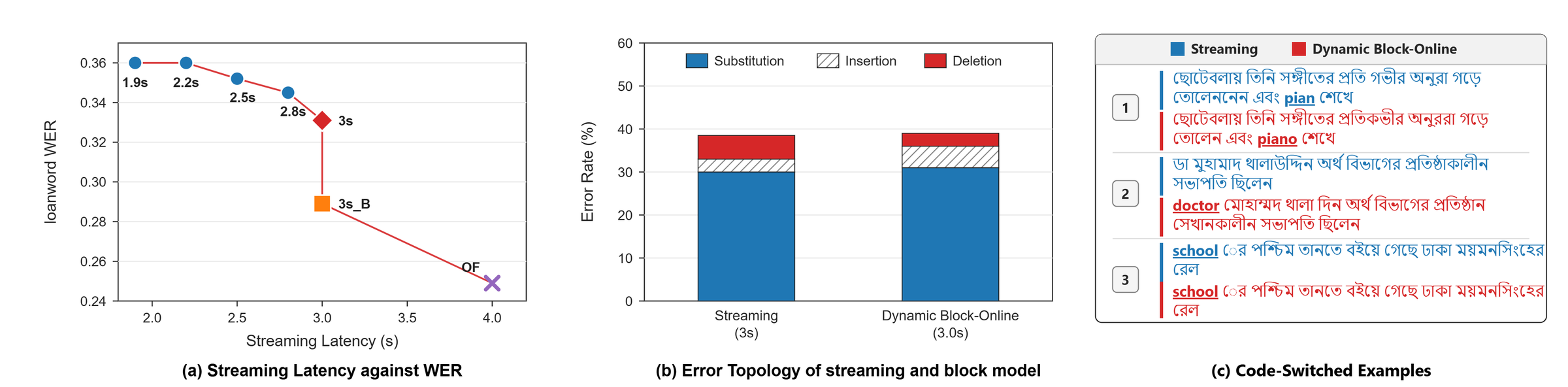} 
    \end{minipage}
    \hfill
    \caption{Illustration of different model outcomes by the streaming, block online, and offline model: (a) Streaming latency against $E_{root}$ of streaming models as circles and diamonds, compared to block online (square) and offline (cross) models. (b) Error topology showcasing the Global WER of the streaming model and the dynamic block-online model (c) Code-Switched generations of dynamic block-online (red) and streaming model (blue).}
    \label{fig:wer_with_table}
\end{figure*}

\subsection{Results and Discussion}
\textbf{Morphological Plateau:}
Table \ref{tab:Results1} reveals that by aggressively extending the Causal Streaming window to $w=2.4$ or $w=3$, we achieve a competitive Global WER of 39.45\% and 37.20\%, matching the Dynamic Block model (38.73\%). However, the CS-WER reveals the structural limitation of the streaming architecture.
The Streaming model saturates at an $E_{root}$ of 0.35 and $E_{morph}$ around $0.42$.
The Dynamic Block model has $E_{root}$ of 0.29 and $E_{morph}$ around $0.35$.
This 6 and 7 point absolute gap in morphological error demonstrates that while large-window streaming can capture general acoustics (lowering Global WER), it fails to resolve the specific dependencies of agglutinative CS. The Causal mask prevents the model from effectively understanding sudden switches to English words and also "binding" the suffix to the root. The Dynamic Block architecture, utilizing bidirectional attention, breaks this plateau, matching the morphological precision of the offline topline. At the same time, it takes many hours of computation and data to adapt the streaming model on the pretrained offline model.
\noindent\textbf{Dynamic Latency Characteristics}
Unlike fixed-window streaming, the latency of our Block-Online system is data-dependent, governed by the speaker's inter-pausal units (IPUs). Since the VAD triggers processing at natural semantic boundaries (pauses $>$ 200ms), the effective latency aligns with the conversational turn-taking rate. In standard conversational speech, phrases typically span 1.5s-2.0s. Consequently, the system rarely hits the hard 3000ms cap during dialogue. The 3-second buffer primarily serves as a fallback for rapid, continuous speech segments with the highest morphological density.

\noindent\textbf{Error Topology and Qualitative Analysis:} Figure~\ref{fig:wer_with_table}b shows error type distribution. The Causal Streaming baseline is dominated by Deletions ($\approx$7\%); the CIF predictor discards suffixes or code-switches. The Dynamic Block-Online architecture has deletions drop to $<$2\% as errors shift mostly to insertions; the globally trained encoder fires at pieces such as suffixes, occasionally over-firing. By leveraging a 3s receptive field and global attention training, the CIF mechanism accumulates sufficient acoustic energy to trigger token emissions for agglutinative markers and switches. Standard WER penalizes both, the block architecture preserves the morphological slot and overall important code-switched semantics, maintaining the integrity required for downstream NLU tasks. While both count equally towards Global WER, Deletion is fatal for semantic understanding in agglutinative languages. By shifting errors from structural deletion to insertion, the Block-Online architecture preserves the grammatical intent, even if the Global WER metric does not fully reflect this semantic recovery. Figure~\ref{fig:wer_with_table}c shows CS examples audios generated through the streaming model (blue) and block online model (red). For first two examples, the streaming model is unable to understand the start of the English loanwords in (1) piano and (2) doctor. This is due to (1) Abrupt cutting of audios and (2) less contextual knowledge due to masked-chunk attention, respectively. 

\noindent\textbf{Impact of Script-Anchored Injection}
The efficacy of our data augmentation strategy is quantified by the Fine-Grained CS-WER metric in Table \ref{tab:Results1}. Models trained on monolingual transcripts (B+E) fail to trigger CS, resulting in near-total error rates as the model transliterates English terms into Bengali script. The application of \textbf{Script-Anchored Loanword Injection} drastically reduces $E_{root}$ in the best configuration. This confirms that the injection strategy successfully transforms the task from simple transcription to discriminative language switching, forcing the model to respect the orthographic boundary between the English root and the Bengali suffix. We see a huge increase in catching loanwords in both offline (87\% $\to$25\%) and online (99\% $\to$ 61\%)

\noindent \textbf{Explicit vs. Implicit Signals}
Explicit LID tags (Table \ref{tab:Results1}) fail to model boundaries: discrete \texttt{<en>} / \texttt{<bn>} tokens destabilize the CIF predictor by interrupting its continuous energy accumulation, disrupting alignment paths. This indicates that Implicit Script Anchoring is better suited for energy-based NAR architectures. Applying weighted loss to LID tokens yields similar performance, suggesting the limitation is architectural rather than optimization-related.

\subsection{Domain Adaptation on Menstrual Health}
We adapt our architecture to a high-perplexity domain: Menstrual and Menopausal Health. This domain features critical, low-frequency English medical entities (e.g., \textit{PCOS, Endometriosis}) within Bangla syntax. Due to data scarcity, we utilize a synthetic-to-real adaptation strategy, generating a domain-specific corpus via TTS.

\noindent \textbf{Results and Semantic Viability:} 
The model initially struggled with specialized vocabulary ($E_{root}=.78$). After fine-tuning on the synthetic subset, the Block-Online model achieved a 27\% WER and an $E_{root}$ of .22 on a noisy, held-out test set—matching the baseline performance on general data. This \textit{Lossless Domain Transfer} suggests that the model’s pre-existing morphological segmentation capabilities allow it to integrate new English roots into Bangla syntax without relearning segmentation mechanics.

\section{Conclusion}
We propose a Dynamic Block-Online ASR for Bangla-English CS that overcomes the structural limitations of causal streaming through VAD-aligned inference, global attention, and Script-Anchored Loanword Injection. The approach improves morphological and semantic accuracy under online latency constraints. CS-WER enables fine-grained diagnosis of switch, root, and morphological errors. Experiments show that conventional streaming models plateau structurally, whereas our method surpasses this ceiling and generalizes. The limitations of this study are that it requires a good VAD model to utilize this methodology. The dataset does not include all possible loanwords, but rather the most common loanwords. Future work will explore adaptive latency control and LLM-assisted loanword injection.


\section{Generative AI Use Disclosure}
Generative AI tools were used solely for writing assistance, language editing, and manuscript polishing. No generative AI tools were used to produce any significant part of the scientific content, methodology, experiments, results, analysis, or conclusions of this manuscript. All authors reviewed the manuscript and take full responsibility for its content.

\bibliographystyle{IEEEtran}
\bibliography{biblio}

\end{document}